\documentclass{article}

\usepackage{PRIMEarxiv}
\usepackage[utf8]{inputenc} 
\usepackage[T1]{fontenc}    
\usepackage{hyperref}       
\usepackage{url}            
\usepackage{booktabs}       
\usepackage{amsfonts}       
\usepackage{nicefrac}       
\usepackage{microtype}      
\usepackage{lipsum}
\usepackage{fancyhdr}       
\usepackage{graphicx}       
\graphicspath{{media/}}     
\usepackage{multirow}
\usepackage{graphicx}
\usepackage{algorithm}
\usepackage{algpseudocode}
\usepackage{amsmath}
\usepackage{amssymb}
\usepackage{amsthm}
\usepackage{bm}
\usepackage{pifont}
\usepackage{colortbl}
\newcommand{\etal}{et al.\ }

\title{DRIVE: Dual-Robustness via Information Variability and Entropic Consistency in Source-Free Unsupervised Domain Adaptation}
\author{
  Ruiqiang Xiao\thanks{The first two authors contributed equally to this work.} \\
  HKUST(GZ) \\
  \texttt{ruiqiangxiao@hkust-gz.edu.cn} \\
   \And
  Songning Lai${}^*$ \\
  HKUST(GZ) \\
  Deep Interdisciplinary Intelligence Lab \\
  \texttt{songninglai@hkust-gz.edu.cn} \\
     \And
  Yijun Yang \\
  HKUST(GZ) \\
  \texttt{yyang018@connect.hkust-gz.edu.cn} \\
       \And
  Jiemin Wu \\
  HKUST(GZ) \\
  Deep Interdisciplinary Intelligence Lab \\
  \texttt{jieminwu@hkust-gz.edu.cn} \\
  \And
  Yutao Yue \\
  HKUST(GZ) \\
  Institute of Deep Perception Technology, JITRI\\
  Deep Interdisciplinary Intelligence Lab\\
\texttt{yutaoyue@hkust-gz.edu.cn}\\
  \And
  Lei Zhu\thanks{Correspondence to Prof. Lei Zhu \{leizhu@hkust-gz.edu.cn\}.}\\
  HKUST(GZ) \\
  \texttt{leizhu@hkust-gz.edu.cn} \\
}

\begin{document}

\maketitle

\begin{abstract}

Adapting machine learning models to new domains without labeled data, especially when source data is inaccessible, is a critical challenge in applications like medical imaging, autonomous driving, and remote sensing. This task, known as \textbf{Source-Free Unsupervised Domain Adaptation (SFUDA)}, involves adapting a pre-trained model to a target domain using only unlabeled target data, which can lead to issues such as overfitting, underfitting, and poor generalization due to domain discrepancies and noise. Existing SFUDA methods often rely on single-model architectures, struggling with uncertainty and variability in the target domain.
To address these challenges, we propose \textbf{\underline{DRIVE}} (\textbf{\underline{D}}ual-\textbf{\underline{R}}obustness through \textbf{\underline{I}}nformation \textbf{\underline{V}}ariability and \textbf{\underline{E}}ntropy), a novel SFUDA framework leveraging a \textbf{dual-model architecture}. The two models, initialized with identical weights, work in parallel to capture diverse target domain characteristics. One model is exposed to perturbations via \textbf{projection gradient descent (PGD)} guided by mutual information, focusing on high-uncertainty regions. We also introduce an \textbf{entropy-aware pseudo-labeling strategy} that adjusts label weights based on prediction uncertainty, ensuring the model focuses on reliable data while avoiding noisy regions.
The adaptation process has two stages: the first aligns the models on stable features using a \textbf{mutual information consistency loss}, and the second dynamically adjusts the perturbation level based on the loss from the first stage, encouraging the model to explore a broader range of the target domain while preserving existing performance. This enhances generalization capabilities and robustness against interference. Evaluations on standard SFUDA benchmarks show that DRIVE consistently outperforms previous methods, delivering improved adaptation accuracy and stability across complex target domains.

\end{abstract}

\vspace{-2mm}
\section{Introduction}
\label{sec:intro}

Adapting machine learning models to new domains where labeled data is unavailable—especially when the original source data cannot be reused—represents a complex yet critical challenge in practical applications across fields like medical imaging \cite{li2023enhancing}, autonomous driving \cite{fang2024source,hegde2023source}, and remote sensing \cite{liu2024source}. This task, known as \textbf{Source-Free Unsupervised Domain Adaptation (SFUDA)}, aims to adapt a pretrained model to a target domain using only unlabeled target data. Unlike traditional domain adaptation methods, which typically require access to both source and target data for knowledge transfer \cite{ganin2015unsupervised, kang2019contrastive}, SFUDA operates under stricter conditions due to privacy, storage, and regulatory constraints that prohibit the reuse of source data. Thus, SFUDA has become a crucial area of research for real-world applications where source data is inaccessible.

A major challenge in SFUDA is the high risk of model overfitting or underfitting in the target domain \cite{yuan2024domain,luo2024crots}, especially when domain discrepancies are large, or when the target data exhibits significant variability and noise \cite{zhao2023source,li2024comprehensive}. Without source data to guide the model, SFUDA methods must rely solely on unlabeled target data, which provides limited and often indirect information about the distributional differences between domains. To address this, most SFUDA techniques rely on self-supervised learning, consistency regularization, or confidence-based approaches to iteratively refine the model's adaptation to the target domain \cite{ding2022source, tian2021vdm, kundu2022balancing}. However, these approaches often use single-model architectures that struggle to generalize effectively when the target domain is noisy or uncertain, leading to brittle performance and suboptimal results \cite{lao2021hypothesis, yang2022attracting, tang2022sclm}.

Existing methods like \textbf{DIFO} (Distilling multimodal Foundation models) \cite{tang2024source}, which leverages CLIP \cite{radford2021learning} and a two-stage distillation process for source-free adaptation, have shown promise. However, DIFO still faces challenges in handling substantial uncertainty or variability in the target domain. Its reliance on a fixed distillation strategy, which combines predictions from CLIP and the target model, often results in suboptimal pseudo-labels, especially when the target data is noisy or uncertain. This introduces a significant challenge for reliable domain transfer, as the model may overfit to noisy data or fail to capture the comprehensive underlying distribution of the target domain.

This limitation of single-model approaches motivates us to explore more robust and adaptive frameworks. While DIFO utilizes a single pretrained model for adaptation, it does not fully capitalize on the potential benefits of using multiple models to handle the inherent variability and uncertainty in the target domain. We hypothesize that incorporating a \textbf{dual-model}\footnote{While there may be ambiguity here, we generalize DIFO as a single model architecture because we treat the overall framework of ViL (CLIP) and target model as a single model.} architecture can address these challenges more effectively, by leveraging complementary strengths of two models working in parallel to promote more stable and transferable adaptations.

To this end, we propose \textbf{\underline{DRIVE}} (\textbf{\underline{D}}ual-\textbf{\underline{R}}obustness through \textbf{\underline{I}}nformation \textbf{\underline{V}}ariability and \textbf{\underline{E}}ntropy), a novel SFUDA framework based on DIFO, designed to overcome these limitations. Our approach builds upon the insights from DIFO but introduces several key modifications to enhance robustness and generalizability. Rather than relying on a fixed, weighted combination of CLIP and the target model outputs, we introduce an \textbf{entropy-based strategy} to determine the weight of pseudo-labels. This entropy-aware method accounts for the uncertainty in the model’s predictions, ensuring that the model prioritizes reliable regions of the target domain while avoiding over-reliance on noisy or uncertain labels.

\begin{figure*}
    \centering
    \includegraphics[width=0.9\linewidth]{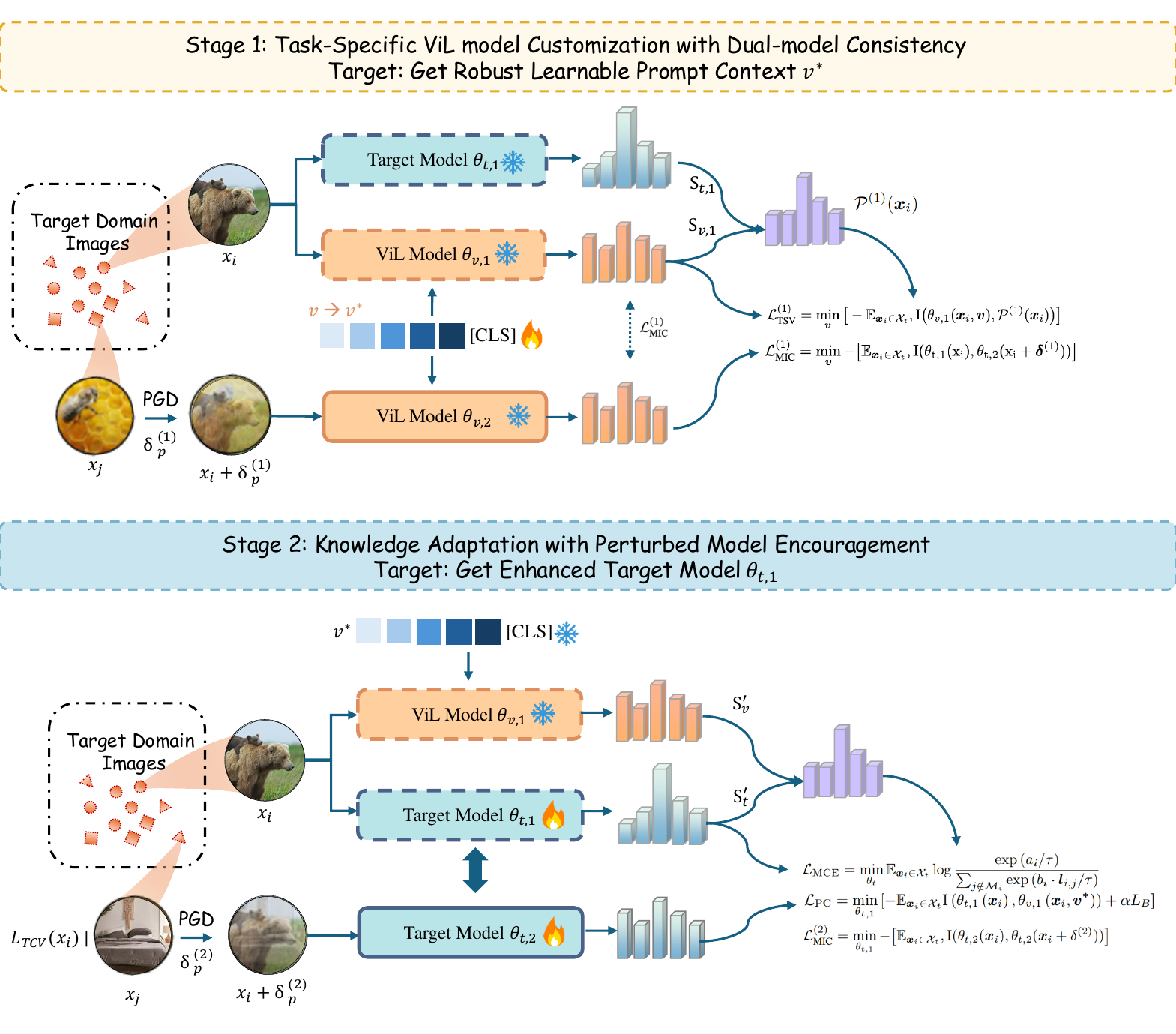}
    \caption{Overview of our DRIVE: it illustrates the two-stage adaptation process of DRIVE, including task-specific ViL model customization and knowledge adaptation with perturbed model encouragement, enhancing robustness and generalization in SFUDA.}
    \label{fig:main}
\end{figure*}
At the core of \textbf{DRIVE} is a dual-model architecture, where two models with identical initial weights operate in parallel throughout the adaptation process. One model is exposed to perturbations generated through \textbf{projection gradient descent (PGD)} \cite{madry2018towards}, guided by mutual information. These perturbations target high-uncertainty regions of the target domain, allowing the model to handle the domain's variability more effectively. This dual-model setup takes advantage of both \textbf{consistency}—encouraging convergence toward stable, transferable features—and \textbf{divergence}—fostering the exploration of diverse characteristics of the target domain. This ensures that the models are not only robust but also capable of capturing the full spectrum of target domain variability.

The adaptation process in \textbf{DRIVE} occurs in two stages. In the first stage, we apply a \textbf{mutual information consistency loss} to align the two models on stable, transferable features despite the presence of perturbations. This phase minimizes the impact of noise, ensuring that the adaptation process is based on reliable features that generalize well across domains. Crucially, the loss function in the first phase also influences the initialization of PGD perturbations in the second phase. By grounding the perturbations in robust, consistent features, the second phase can more effectively explore the target domain's variability, leading to more effective adaptation. This interdependence between the stages ensures that the exploration of the target domain in the second phase is built upon a stable foundation, leading to more effective adaptation and preventing overfitting to noisy data.

In the second stage, we introduce a \textbf{mutual information divergence loss}, which encourages the models to explore complementary aspects of the target data. This phase mitigates the risk of overfitting and enables the models to learn complementary information, enhancing the overall robustness and adaptability of the system. The interplay between the two phases ensures that the first stage's stable learning process guides the more exploratory second phase, leading to a seamless and effective adaptation trajectory.

Our approach offers several key contributions to the SFUDA landscape:

\noindent\underline{\textbf{(i)}} \textbf{Dual-model Architecture}: We introduce a dual-model framework that promotes both consistency (for stable feature alignment) and divergence (for exploring complementary aspects of the target domain), improving the robustness and adaptability of the model in the target domain.

\noindent\underline{\textbf{(i)}} \textbf{Entropy-Aware Pseudo-Labeling}: We propose an entropy-based strategy for pseudo-label weighting, ensuring that the model focuses on reliable regions of the target domain while mitigating the risks of overfitting to uncertain or noisy data.

\noindent\underline{\textbf{(iii)}} \textbf{Dynamic Perturbation Adjustment for Enhanced Exploration}: We propose a two-phase adaptation process, where the first stage aligns models on stable features, and the second stage leverages the first-stage loss to guide the initialization of PGD perturbations, enhancing the exploration of high-uncertainty regions in the target domain.

\section{Related Work}
\label{sec:rel}


\paragraph{Source-Free Unsupervised Domain Adaptation. } 
SFUDA seeks to adapt models to a new, unlabeled target domain without access to the original source data, often restricted due to privacy or storage constraints. Recent SFUDA methods have emphasized self-supervised learning, leveraging pseudo-labeling and entropy minimization to infer target domain structures~\cite{liang2020we, yang2021nrc, litrico2023guiding}. Consistency regularization techniques promote model stability across augmented target data, effectively reducing uncertainties in the target domain~\cite{ge2022domain, lee2022confidence, lai2023padclip,tang2022sclm}. Additionally, approaches such as CPGA~\cite{qiu2021source} and BAIT~\cite{yang2020unsupervised} employ contrastive learning frameworks to align samples with category-wise prototypes. NRC~\cite{yang2021exploiting} and LSC-SDA~\cite{yang2021generalized} propagate categorical semantics through neighborhood or cluster structures in the feature space.
Xia \etal~\cite{xia2021adaptive} focus on the disagreements between target data and the source model.
Litrico \etal~\cite{litrico2023guiding} leverage a loss reweighting strategy that brings robustness against the noise that inevitably affects the pseudo-labels.
DIFO~\cite{tang2024source} explores the integration of heterogeneous knowledge sources from vision-language models.

Despite these advancements, most SFUDA methods rely on single-model frameworks that are susceptible to noise and uncertainty in target data, especially when domain gaps are substantial. Our proposed method addresses this limitation by introducing a dual-model structure, incorporating both consistency and divergence regularizations. This approach leverages mutual information to enhance robustness across diverse target conditions.

\paragraph{Mutual Information.} 
Mutual information (MI) has emerged as a valuable tool in domain adaptation, enabling models to capture and retain essential information across domains without requiring direct alignment data. MI-based alignment techniques aim to maximize shared information between model representations and target domain features, thereby enhancing adaptation robustness and the quality of extracted features. Approaches such as those by Peng \etal~\cite{peng2019moment} and Singha \etal~\cite{singha2023ad} focus on MI-based metrics to strengthen domain-invariant feature extraction. In vision-language models (VLMs), MI-based techniques facilitate the alignment of cross-modal information by maximizing dependency between visual and language representations, ensuring robust cross-modal grounding. For instance, Ma \etal~\cite{ma2023liv} introduce mutual information contrastive learning to align vision-language representations, grounding language specifications and rewarding learning from action-free videos with text annotations. Kim \etal~\cite{kim2022mutual} propose negative Gaussian cross-mutual information, coined as Mutual Information Divergence, using CLIP features as a unified metric.
DIFO~\cite{tang2024source} customize VLMs by maximizing the mutual information with the
target model in a prompt learning manner.

Beyond facilitating domain adaptation and cross-modal semantic alignment, MI also enables model-specific perturbations that enhance adaptability and responsiveness to target-specific uncertainties. Our approach leverages MI both as a consistency mechanism in the initial phase and as a divergence mechanism in the later stage, enabling the framework to harness both alignment and diversity during model learning. This dual-phase application of MI ensures robust feature alignment while promoting model flexibility across diverse target conditions.

\section{Method}
\label{sec:method}




\subsection{Problem Statement and Overview}

\noindent \textbf{Problem Statement.} 
This work addresses the Source-Free Unsupervised Domain Adaptation (SFUDA) problem, where the objective is to adapt a pre-trained source model $\theta_s$ to an unlabeled target domain $\mathcal{X}_t$ without direct access to the labeled source data $\mathcal{X}_s$, often due to privacy or storage constraints. Formally, we consider two domains that share $C$ common classes: a labeled source domain $\mathcal{X}_s$ and an unlabeled target domain $\mathcal{X}_t$ comprising $n$ target samples $\{\boldsymbol{x}_{i}\}_{i=1}^{n}$, with unknown true labels $\mathcal{Y}_t = \{y_{i}\}_{i=1}^{n}$ in the target domain. The goal is to adapt $\theta_s$ to a target model $\theta_t: \mathcal{X}_t \to \mathcal{Y}_t$ using only $\mathcal{X}_t$ and $\theta_s$.

\noindent \textbf{Overview of DRIVE Framework.} The DRIVE framework employs a dual-model, two-stage approach with perturbation-induced robustness. It leverages Projected Gradient Descent (PGD) noise perturbations, where the loss computed in Stage 1 influences the initialization of the perturbation in Stage 2. Both stages utilize an entropy-based pseudo-labeling strategy. Details of the two stages are outlined below.

\subsection{Stage 1: Task-Specific ViL Model Customization with Dual-Model Consistency} 
\label{sec:tsc}

In the first stage, DRIVE employs prompt learning to customize the visual-language (ViL) model $\theta_v$ by aligning its predictions with those of the frozen target model $\theta_t$ initialized with $\theta_s$. This alignment is achieved through a dual-model approach, where Model 1 ($\theta_{v,1}, \theta_{t,1}$) processes clean target samples, while Model 2 ($\theta_{v,2}, \theta_{t,2}$) is perturbed using Projected Gradient Descent (PGD). 

A PGD perturbation is defined as an adversarial adjustment to $\boldsymbol{x}$, designed to maximize a loss function $L(\theta_t(\boldsymbol{x}), y)$ with respect to $\boldsymbol{x}$ while keeping the perturbed sample within an $\epsilon$-ball around the original input $\boldsymbol{x}$. Formally, a PGD perturbation is given by:
\begin{equation}
    \boldsymbol{x}^{\text{PGD}} = \text{Proj}_{\mathcal{B}(\boldsymbol{x}, \epsilon)} \left( \boldsymbol{x} + \alpha \, \nabla_{\boldsymbol{x}} L(\theta_t(\boldsymbol{x}), y) \right),
\end{equation}

where $\text{Proj}_{\mathcal{B}(\boldsymbol{x}, \epsilon)}$ denotes projection onto the $\epsilon$-ball around $\boldsymbol{x}$, and $\alpha$ is a step size.

The perturbations in Model 2, guided by mutual information, explore meaningful variations in the target domain, while the weights of Models 1 and 2 are shared. Importantly, $\theta_{t,1}$ and $\theta_{t,2}$ remain frozen throughout Stage 1.

\noindent \textbf{Entropy-Aware Predictor.} At this stage, pseudo-labels for the task are generated by combining the categorical distribution outputs of $\theta_{t,1}(\boldsymbol{x}_i)$ and $\theta_{v,1}(\boldsymbol{x}_i, \boldsymbol{v})$ with the learnable prompt context $\boldsymbol{v}$, weighted by their respective entropies. The pseudo-label for Stage 1 is defined as follows:
\begin{align}
    \label{eqn:s1_pl}
    &\mathcal{P}^{(1)}(\boldsymbol{x}_i) = \frac{\mathcal{S}_{v,1}(\boldsymbol{x}_i)}{\mathcal{S}_{v,1}(\boldsymbol{x}_i)+\mathcal{S}_{t,1}(\boldsymbol{x}_i) + \lambda}\cdot\theta_{t,1}(\boldsymbol{x}_i)\\ \nonumber
    &+ \frac{\mathcal{S}_{t,1}(\boldsymbol{x}_i) + \lambda}{\mathcal{S}_{v,1}(\boldsymbol{x}_i)+\mathcal{S}_{t,1}(\boldsymbol{x}_i) + \lambda}\cdot\theta_{v,1}(\boldsymbol{x}_i, \boldsymbol{v})
\end{align}

Here, $\mathcal{S}_{v,1}(\boldsymbol{x}_i)$ and $\mathcal{S}_{t,1}(\boldsymbol{x}_i)$ represent the entropy of the categorical distribution output from $\theta_{v,1}(\boldsymbol{x}_i, \boldsymbol{v})$ and $\theta_{t,1}(\boldsymbol{x}_i)$, respectively. $\lambda$ is a prior bias used to adjust the model's confidence in the output of the ViL model.

\noindent \textbf{Task-Specific ViL Loss.} To align the predictions of $\theta_{v,1}(\boldsymbol{x}_i)$ with $\mathcal{P}^{(1)}(\boldsymbol{x}_i)$, we apply a mutual information-based consistency loss. The loss is formulated as:
\begin{equation}
    \label{eqn:loss_tsv}
        \mathcal{L}_{\text{TSV}}^{(1)} = \min_{\boldsymbol{v}} \big[-\mathbb{E}_{\boldsymbol{x}_i \in \mathcal{X}_t} , {\rm{I}} \big(\theta_{v,1}(\boldsymbol{x}_i, \boldsymbol{v}), \mathcal{P}^{(1)}(\boldsymbol{x}_i)\big)\big]
\end{equation}
where ${\rm{I}}(\cdot, \cdot)$ denotes the mutual information between the two predictions.

\noindent \textbf{Mutual Information Consistency Loss for Stage 1.} In this step, we apply a mutual information-based consistency loss to align predictions from both the clean and perturbed models. Let the predictions from Model 2 be denoted by $\theta_{v,2}(\boldsymbol{x}_i + \delta^{(1)}, \boldsymbol{v})$ for a target sample $\boldsymbol{x}_i$. The mutual information-based consistency loss is given by:
\begin{align}
    \label{eqn:loss_mic1}
\mathcal{L}_{\text{MIC}}^{(1)} = \min_{\boldsymbol{v}} -\big[\mathbb{E}_{\boldsymbol{x}_i \in \mathcal{X}_t} , \rm{I} (\theta_{t,1}(\boldsymbol{x}_i),\theta_{t,2}(\boldsymbol{x}_i + \boldsymbol{\delta}^{(1)}))\big]
\end{align}

This loss encourages Model 1 to generalize through the learnable text prompt context $\boldsymbol{v}$ in a way that remains consistent with the perturbed Model 2, thus aligning both models' outputs under varying target domain conditions.

The optimization process follows the PGD methodology \cite{madry2018towards}, where iterative updates to the first stage perturbation $\boldsymbol{\delta}^{(1)}$ are performed. At the $p$-th iteration, the current perturbation $\boldsymbol{\delta}_{p-1}^{*(1)}$ is updated as follows:
\begin{align}
    \label{eqn:pgd_delta_stage1}
&\boldsymbol{\delta}_p^{(1)} = \boldsymbol{\delta}_{p-1}^{*(1)} +  \frac{\gamma_p }{|A_{p-1}|}\sum_{x\in A_{p-1}} \nabla_{\boldsymbol{\delta}_{p-1}^{*(1)}} \nonumber\\
&[ -\mathbb{E}_{\boldsymbol{x}_i \in \mathcal{X}_t} , {\rm{I}} \big(\theta_{v,1}(\boldsymbol{x}_i, \boldsymbol{v}), \mathcal{P}^{(1)}(\boldsymbol{x}_i)\big)]
\end{align}
where $\boldsymbol{\delta}_p^{*(1)} =\arg\min_{||\boldsymbol{\delta}^{(1)}|| \leq R} ||\boldsymbol{\delta}^{(1)} - \boldsymbol{\delta}_p^{(1)} ||$ and $A_{p-1}$ denotes a batch of samples, $\gamma_p$ is the step size parameter for PGD, and $R$ is the norm bound for the perturbation. Specifically, the initialization of $\boldsymbol{\delta}_p^{(1)} - \boldsymbol{\delta}^{(1)} = \textbf{Random}(\boldsymbol{x}_j)$ is a random input from this batch of samples $A_{p-1}$. Once $\boldsymbol{\delta}_P^{(1)}$ is obtained after $P$ iterations, we update $\boldsymbol{v}$ to $\boldsymbol{v}^*$ using batched gradients.

\subsection{Stage 2: Knowledge Adaptation with Perturbed Model Encouragement} 
\label{sec:mi-based-consistency}

In this stage, we extend the dual-model approach from Stage 1 to construct PGD perturbations, driving the target model $\theta_{t, 1}$ to perform more extensive exploration of the target domain based on its overall consistency loss from Stage 1.

\noindent \textbf{Dynamic Perturbation Adjustment for Enhanced Exploration.} To ensure that the model performs extensive exploration in the target domain while maintaining prediction consistency for familiar samples, DRIVE introduces a dynamic perturbation adjustment mechanism in Stage 2. This mechanism dynamically adjusts the initialization noise magnitude of the PGD perturbations based on the consistency losses from Stage 1. The perturbation noise magnitude $\eta$ for Stage 2 is defined as:
\begin{equation}
    \label{eqn:noise_adjustment}
    \eta \propto \mathcal{L}_{\text{TSV}}^{(1)} + \beta \mathcal{L}_{\text{MIC}}^{(1)}
\end{equation}

The optimization process for these perturbations follows the PGD methodology \cite{madry2018towards}. At the $p$-th iteration, the current perturbation $\boldsymbol{\delta}_{p-1}^{*(2)}$ is updated similarly to the process described earlier, but with specific adjustments for the iteration and initialization. Specifically, we compute $\mathcal{P}^{(2)}(\boldsymbol{x}_i)$ as the pseudo-label for Stage 2 computed based on Eq.~\ref{eqn:s1_pl} and the target-domain-customized context prompt embedding $\boldsymbol{v^*}$. In addition, the initialization of $\boldsymbol{\delta}_p^{(2)} - \boldsymbol{\delta}_0^{(2)} = \eta \cdot \textbf{Random}(\boldsymbol{x}_j)$ is a random input from this batch of samples $A_{p-1}$, scaled by $\eta$.

when the target model achieves high consistency for a target domain sample $\boldsymbol{x}_i$ in Stage 1, it indicates that the model has learned the label correspondence between this sample and the source domain samples. Choosing a smaller initialization noise helps retain this learned stable mapping in subsequent training. Conversely, if the target model's predictions for the same target domain sample $\boldsymbol{x}_i$ are inconsistent with the existing ViL prior knowledge and the predictions after PGD perturbations in Stage 1, it suggests that the distribution space represented by this sample remains highly uncertain for the current target model. Increasing the magnitude of the initialization noise aids the model in performing more extensive exploration within the neighborhood of this sample.

\noindent \textbf{Mutual Information Consistency Loss for Stage 2.} As in Stage 1, to ensure that the perturbed and unperturbed models align under a wide range of perturbations, we use mutual information-based consistency losses. The mutual information consistency loss for Stage 2 is formulated as:
\begin{align}
    \label{eqn:loss_mic2}
    \mathcal{L}_{\text{MIC}}^{(2)} = \min_{\theta_{t,1}} -\big[\mathbb{E}_{\boldsymbol{x}_i \in \mathcal{X}_t} , {\rm{I}}(\theta_{t,2}(\boldsymbol{x}_i), \theta_{t,2}(\boldsymbol{x}_i + \delta^{(2)}))\big]
\end{align}
This loss encourages Model 1 to generalize in a manner that remains consistent with the perturbed Model 2, aligning both models' outputs under varying target domain conditions.

\noindent \textbf{Predictive Consistency and Category Attention Calibration.} To ensure knowledge adaptation and improve model performance, we incorporate two key components following the DIFO method: predictive consistency loss and category attention calibration.

First, the predictive consistency loss ensures that the target model's predictions remain consistent with those of the ViL model. This loss is defined as:
\begin{small} 
\begin{equation}
    \label{eqn:loss_pc}
    \begin{aligned}
        \mathcal{L}_{\rm{PC}} = \min_{\theta_{t,1}} \left[- \mathbb{E}_{{\boldsymbol{x}}_{i} \in {\mathcal{X}_t}} {{\rm{I}}}\left(\theta_{t,1}\left({\boldsymbol{x}}_{i}\right), \theta_{v,1} \left(\boldsymbol{x}_{i}, \boldsymbol{v^*}\right)\right) + \alpha L_B \right]
    \end{aligned}
\end{equation}
\end{small} 
where the category balance term $L_B = \text{KL}(\theta_{v,1} (\boldsymbol{x}_{i}) \| \boldsymbol{\frac{1}{C}})$ ensures the predicted label distribution matches the uniform distribution $\boldsymbol{\frac{1}{C}}$.

Second, we employ category attention calibration to regularize the model's predictions using pseudo-labels. Specifically, we identify the top-$N$ most probable categories using $\mathcal{P}_{s,1}(\boldsymbol{x}_i)$. The indices of these categories are denoted by $\mathcal{M}_i=\{m_k\}_{k=1}^{N}$. The regularization loss is defined as:
\begin{equation}\label{eqn:loss_mce} 
    \begin{split} 
    \mathcal{L}_{\rm{MCE}} &= \min_{\theta_t} \mathbb{E}_{{\boldsymbol{x}}_{i} \in {\mathcal{X}_t}}  
    \log \frac{\exp\left( a_i / \tau \right)}{\sum_{\substack{j \notin \mathcal{M}_i}} \exp{\left( b_i \cdot \boldsymbol{l}_{i,j} / \tau \right)}}\\
    a_i &= \prod\limits_{k=1}^{N} {\boldsymbol{l}_{i,m_k}}, \quad
    b_i = \sum\limits_{k=1}^{N} {\boldsymbol{l}_{i,m_k}}
    \end{split}
\end{equation}
where $\boldsymbol{l}_{i, j}$ denotes the $j$-th element of the logit vector $\boldsymbol{l}_{i}$ and $\tau$ is the temperature parameter.

Together, these mechanisms ensure that the target model not only adapts to the target domain but also explores new and challenging examples, thereby improving its generalization and robustness.

\subsection{Training Procedure}
The training process for DRIVE iterates between Stage 1 (ViL model customization) and Stage 2 (knowledge adaptation), progressively adapting $\theta_{t,1}$ and $\theta_{t,2}$ by leveraging both clean and perturbed inputs. In each epoch, Stage 1 optimizes prompt $\boldsymbol{v}$ with $\mathcal{L}_{\text{total}}^{(1)} = \mathcal{L}_{\text{TSV}}^{(1)} + \beta \mathcal{L}_{\text{MIC}}^{(1)}$, while Stage 2 adapts $\theta_{t,1}$ using: 
\begin{equation} 
\label{eqn:s2} 
\mathcal{L}_{\text{total}}^{(2)} = \mathcal{L}_{\text{MCE}}+\xi_1 \mathcal{L}_{\text{PC}} + \xi_2 \mathcal{L}_{\text{MIC}}^{(2)}
\end{equation} 

Here, $\xi_1$ and $\xi_2$ are hyperparameters that control the weights of the respective loss components, $\mathcal{L}_{\text{PC}}$ and $\mathcal{L}_{\text{MIC}}^{(2)}$. The details of this procedure are summarized in Algorithm \ref{alg}.

\begin{algorithm}
\caption{Training Algorithm for DRIVE Framework}
\label{alg}
\begin{algorithmic}[1]
\State \textbf{Input:} Target domain $\mathcal{X}_t$, ViL model $\theta_v$, target model $\theta_t$, perturbation magnitude $\eta_0$, balancing factor $\beta$, $\xi_1$, $\xi_2$
\State \textbf{Output:} Adapted target model $\theta_t$
\State Initialize target model $\theta_t$ and ViL model $\theta_v$ with pre-trained weights
\For{each epoch} 
    \State \textbf{Stage 1: ViL Model Customization}
    \For{each sample $\boldsymbol{x}_i \in \mathcal{X}_t$}
        \State Sample perturbation $\boldsymbol{\delta}$ with magnitude $\eta_0$
        \State Compute clean prediction from target model: $\theta_{t,1}(\boldsymbol{x}_i)$
        \State Compute clean prediction from ViL model: $\theta_{v,1}(\boldsymbol{x}_i, \boldsymbol{v})$
        \State Compute perturbed prediction from ViL model: $\theta_{v,2}(\boldsymbol{x}_i + \boldsymbol{\delta}, \boldsymbol{v})$
        \State Compute pseudo-label using entropy-aware predictor: $\mathcal{P}^{(1)}(\boldsymbol{x}_i)$
        \State Calculate mutual information consistency loss: $\mathcal{L}_{\text{total}}^{(1)} = \mathcal{L}_{\text{TSV}}^{(1)} + \beta \mathcal{L}_{\text{MIC}}^{(1)}$
        \State Update prompt embedding $\boldsymbol{v}$ using $\mathcal{L}_{\text{total}}^{(1)}$
    \EndFor

    \State \textbf{Stage 2: Knowledge Adaptation}
    \For{each sample $\boldsymbol{x}_i \in \mathcal{X}_t$}
        \State Compute perturbation: $\boldsymbol{\delta}$ with adaptive dynamical magnitude $\eta$ and $\mathcal{L}_{\text{total}}^{(1)}$
        \State Compute pseudo-label using entropy-aware predictor: $\mathcal{P}^{(2)}(\boldsymbol{x}_i)$
        \State Compute mutual information-divergence loss: $\mathcal{L}_{\text{MID}}$
        \State Compute predictive consistency loss: $ \mathcal{L}_{\text{PC}}$
         \State Compute category attention calibration loss $\mathcal{L}_{\text{MCE}}$
        \State Update target model $\theta_t$ using the combined loss:
        \Statex \quad $\mathcal{L}_{\text{total}}^{(2)} = \mathcal{L}_{\text{MCE}}+\xi_1 \mathcal{L}_{\text{PC}} + \xi_2 \mathcal{L}_{\text{MIC}}^{(2)}$
    \EndFor
\EndFor
\end{algorithmic}
\end{algorithm}

\section{Experiments}
\label{sec:experiments}

\subsection{Experimental Setup}

\noindent \textbf{Datasets: }We evaluate our proposed framework, \textbf{DRIVE}, on four benchmark datasets for domain adaptation: \textbf{Office-31} \cite{saenko2010adapting}, \textbf{Office-Home} \cite{venkateswara2017deep}, and \textbf{DomainNet-126} \cite{peng2019moment}. These datasets offer varying levels of complexity, from small-scale (Office-31) to large-scale (Ofiice-Home, DomainNet-126), ensuring a comprehensive assessment across different domain shifts and challenges. Dataset details are provided in the Supplementary Materials.

\noindent \textbf{Baselines: }We compare DRIVE with leading SFUDA methods across three categories:\textbf{(i) Source model: } Baselines include \textit{Source} (the source-only model). \textbf{(ii) Multimodal UDA Methods: }We include DAPL~\cite{ge2022domain}, PADCLIP~\cite{lai2023padclip}, ADCLIP~\cite{singha2023ad}, and DIFO~\cite{tang2024source}, which leverage multimodal models for domain adaptation. \textbf{(iii) SFUDA Methods: }Key SFUDA methods include SHOT~\cite{liang2020we}, NRC~\cite{yang2021nrc}, and TPDS~\cite{tang2023source}, which address domain shifts without requiring source data during adaptation.

\noindent \textbf{Implementation Details: }WWe present DRIVE-C-B32, which utilizes the ViT-B/32 backbone, and is designed to be broadly similar to DIFO-C-B2 (also using ViT-B/32). Experiments follow the same settings as previous baselines for fair comparison, with further details in the Supplementary Materials.

\subsection{Results on Closed-set SFUDA}
On the Office-31 dataset(Table \ref{tab:office31}), DRIVE achieves a mean accuracy of 92.7\%, significantly outperforming state-of-the-art methods. Specifically, in the challenging tasks such as A → D (97.4\%) and A → W (95.7\%), DRIVE demonstrates robustness and adaptability. The entropy-aware pseudo-labeling strategy ensures that the model focuses on reliable regions, reducing overfitting to noisy data. This is particularly important in tasks where the target domain has high variability and noise.
\begin{table*}[htbp]
  \centering
  \caption{Closed-set SFDA on \textbf{Office-31} (\%).}
  \resizebox{0.8 \columnwidth}{!}{%
    \begin{tabular}{ll|ccccccc}
      \toprule
      Method & Venue & A → D & A → W & D → A & D → W & W → A & W → D & Mean \\
      \midrule
      Source & --     & 79.1  & 76.6  & 59.9  & 95.5  & 61.4  & 98.8  & 78.6  \\
      SHOT  & ICML20 & 93.7  & 91.1  & 74.2  & 98.2  & 74.6  & \textbf{100.0} & 88.6  \\
      NRC   & NIPS21 & 96.0  & 90.8  & 75.3  & \textbf{99.0} & 75.0  & \textbf{100.0} & 89.4  \\
      GKD   & IROS21 & 94.6  & 91.6  & 75.1  & 98.7  & 75.1  & \textbf{100.0} & 89.2  \\
      HCL   & NIPS21 & 94.7  & 92.5  & 75.9  & 98.2  & 77.7  & \textbf{100.0} & 89.8  \\
      AaD   & NIPS22 & 96.4  & 92.1  & 75.0  & 99.1  & 76.5  & \textbf{100.0} & 89.9  \\
      AdaCon  & CVPR22 & 87.7  & 83.1  & 73.7  & 91.3  & 77.6  & 72.8  & 81.0  \\
      CoWA  & ICML22 & 94.4  & 95.2  & 76.2  & 98.5  & 77.6  & 99.8  & 90.3  \\
      SCLM  & NN22  & 95.8  & 90.0  & 75.5  & 98.9  & 75.5  & 99.8  & 89.4  \\
      ELR   & ICLR23 & 93.8  & 93.3  & 76.2  & 98.0  & 76.9  & \textbf{100.0} & 89.6  \\
      PLUE  & CVPR23 & 89.2  & 88.4  & 72.8  & 97.1  & 69.6  & 97.9  & 85.8  \\
      TPDS  & IJCV23 & 97.1  & 94.5  & 75.7  & 98.7  & 75.5  & 99.8  & 90.2  \\
      DIFO-C-B32 & CVPR24 & 96.2  & 95.0  & 83.1  & 96.0  & 83.0  & 99.0  & 92.0  \\
      \rowcolor[rgb]{ .91,  .91,  .91} \textbf{DRIVE(Ours)} & --     & \textbf{97.4} & \textbf{96.0} & \textbf{83.5} & 96.9  & \textbf{83.1} & 99.4  & \textbf{92.7} \\
      \bottomrule
    \end{tabular}%
  }
  \label{tab:office31}%
\end{table*}
For the Office-Home dataset(Table \ref{tab:office-home}), DRIVE achieves a mean accuracy of 83.6\%. Notable improvements are observed in tasks such as Ar → Cl (72.4\%) and Cl → Pr (90.7\%), which are known for their high uncertainty. The dynamic perturbation adjustment mechanism enhances the exploration of high-uncertainty regions, leading to better feature alignment and generalization. This is crucial for tasks where the target domain has significant variations, and the model needs to adapt effectively to these changes.
\begin{table*}[htbp]
  \centering
  \caption{Closed-set SFDA on Office-home (\%). SF and M means source-free and multimodal, respectively.}
  \resizebox{1 \columnwidth}{!}{%
    \begin{tabular}{ll|c|c|ccccccccccccc}
    \toprule
    \multirow{2}[2]{*}{Method} & \multirow{2}[2]{*}{Venue} & \multirow{2}[2]{*}{\textbf{SF}} & \multirow{2}[2]{*}{\textbf{M}} & \multicolumn{13}{c}{\textbf{Office-home}} \\
          &       &       &       & Ar → Cl & Ar → Pr & Ar → Rw & Cl → Ar & Cl → Pr & Cl → Rw & Pr → Ar & Pr → Cl & Pr → Rw & Rw → Ar & Rw→ Cl & Rw → Pr & Mean \\
    \midrule
    Source & --     & -     & -     & 43.7  & 67.0  & 73.9  & 49.9  & 60.1  & 62.5  & 51.7  & 40.9  & 72.6  & 64.2  & 46.3  & 78.1  & 59.2  \\
    \midrule
    DAPL-RN & TNNLS23 & \ding{53}     & \checkmark     & 54.1  & 84.3  & 84.8  & 74.4  & 83.7  & 85.0  & 74.5  & 54.6  & 84.8  & 75.2  & 54.7  & 83.8  & 74.5  \\
    PADCLIP-RN & ICCV23 & \ding{53}     & \checkmark     & 57.5  & 84.0  & 83.8  & 77.8  & 85.5  & 84.7  & 76.3  & 59.2  & 85.4  & 78.1  & 60.2  & 86.7  & 76.6  \\
    ADCLIP-RN & ICCVW23 & \ding{53}     & \checkmark     & 55.4  & 85.2  & 85.6  & 76.1  & 85.8  & 86.2  & 76.7  & 56.1  & 85.4  & 76.8  & 56.1  & 85.5  & 75.9  \\
    \midrule
    SHOT  & ICML20 & \checkmark     & \ding{53}     & 56.7  & 77.9  & 80.6  & 68.0  & 78.0  & 79.4  & 67.9  & 54.5  & 82.3  & 74.2  & 58.6  & 84.5  & 71.9  \\
    NRC   & NIPS21  & \checkmark     & \ding{53}     & 57.7  & 80.3  & 82.0  & 68.1  & 79.8  & 78.6  & 65.3  & 56.4  & 83.0  & 71.0  & 58.6  & 85.6  & 72.2  \\
    GKD   & IROS21  & \checkmark     & \ding{53}     & 56.5  & 78.2  & 81.8  & 68.7  & 78.9  & 79.1  & 67.6  & 54.8  & 82.6  & 74.4  & 58.5  & 84.8  & 72.2  \\
    AaD   & NIPS22  & \checkmark     & \ding{53}     & 47.2  & 75.1  & 75.5  & 60.7  & 73.3  & 73.2  & 60.2  & 45.2  & 76.6  & 65.6  & 48.3  & 79.1  & 65.0  \\
    AdaCon  & CVPR22  & \checkmark     & \ding{53}     & 56.9  & 78.4  & 81.0  & 69.1  & 80.0  & 79.9  & 67.7  & 57.2  & 82.4  & 72.8  & 60.5  & 84.5  & 72.5  \\
    CoWA  & ICML22  & \checkmark     & \ding{53}     & 56.9  & 78.4  & 81.0  & 69.1  & 80.0  & 79.9  & 67.7  & 57.2  & 82.4  & 72.8  & 60.5  & 84.5  & 72.5  \\
    SCLM  & NN22  & \checkmark     & \ding{53}     & 58.2  & 80.3  & 81.5  & 69.3  & 79.0  & 80.7  & 69.0  & 56.8  & 82.7  & 74.7  & 60.6  & 85.0  & 73.0  \\
    ELR   & ICLR23 & \checkmark     & \ding{53}     & 58.4  & 78.7  & 81.5  & 69.2  & 79.5  & 79.3  & 66.3  & 58.0  & 82.6  & 73.4  & 59.8  & 85.1  & 72.6  \\
    PLUE  & CVPR23  & \checkmark     & \ding{53}     & 49.1  & 73.5  & 78.2  & 62.9  & 73.5  & 74.5  & 62.2  & 48.3  & 78.6  & 68.6  & 51.8  & 81.5  & 66.9  \\
    TPDS  & IJCV23  & \checkmark     & \ding{53}     & 59.3  & 80.3  & 82.1  & 70.6  & 79.4  & 80.9  & 69.8  & 56.8  & 82.1  & 74.5  & 61.2  & 85.3  & 73.5  \\
    DIFO-C-B32 & CVPR24 & \checkmark     & \checkmark     & 70.6  & \textbf{90.6 } & 88.8  & \textbf{82.5 } & 90.6  & 88.8  & \textbf{80.9 } & 70.1  & 88.9  & 83.4  & 70.5  & 91.2  & 83.1  \\
    \rowcolor[rgb]{ .91,  .91,  .91} \textbf{DRIVE(Ours)} & --     & \textbf{\checkmark} & \textbf{\checkmark} & \textbf{72.4 } & 89.9  & \textbf{89.2 } & 81.8  & \textbf{90.7 } & \textbf{89.3 } & \textbf{80.9 } & \textbf{71.2 } & \textbf{90.0 } & \textbf{83.6 } & \textbf{73.2 } & \textbf{91.5 } & \textbf{83.6 } \\
    \bottomrule
    \end{tabular}%
  }
  \label{tab:office-home}%
\end{table*}%

On the DomainNet-126 dataset(Table \ref{tab:domainnet126}), DRIVE achieves a mean accuracy of 80.6\%, outperforming other methods. In tasks such as C → R (88.1\%) and R → C (81.0\%), DRIVE shows strong performance, indicating its ability to handle large domain shifts. The dual-model architecture and mutual information-driven consistency loss ensure stable feature alignment, even in complex target domains. This robust performance across multiple datasets and tasks highlights the effectiveness of DRIVE in addressing the challenges of closed-set SFUDA.

These results substantiate DRIVE’s capability to enhance cross-domain performance in closed-set SFUDA settings, benefiting from its dual-model architecture and mutual information-driven consistency mechanism.

\begin{table*}[htbp]
  \centering
  \caption{Closed-set SFDA on DomainNet-126 (\%). SF and M means source-free and multimodal, respectively.}
  \resizebox{0.8 \columnwidth}{!}{%
    \begin{tabular}{ll|c|c|cccccc}
    \toprule
    Method & Venue & \textbf{SF} & \textbf{M} & C→R   & R→C   & R→P   & R→S   & S→C   & Mean \\
    \midrule
    Source & -     & -     & -     & 59.8  & 55.3  & 62.7  & 46.4  & 55.1  & 55.9  \\
    \midrule
    DAPL-RN & TNNLS23 & \ding{53}     & \checkmark     & 87.6  & 73.2  & 72.4  & 66.2  & 73.8  & 74.6  \\
    ADCLIP-RN & ICCVW23 & \ding{53}     & \checkmark     & 88.1  & 73.6  & 73.0  & 68.4  & 72.3  & 75.1  \\
    \midrule
    SHOT  & ICML20 & \checkmark     & \ding{53}     & 78.2  & 67.7  & 67.6  & 57.8  & 70.2  & 68.3  \\
    NRC   & NIPS21  & \checkmark     & \ding{53}     & 77.1  & 64.7  & 69.4  & 58.7  & 69.4  & 67.9  \\
    GKD   & IROS21  & \checkmark     & \ding{53}     & 77.4  & 68.3  & 68.4  & 59.5  & 71.5  & 69.0  \\
    AdaCon  & CVPR22  & \checkmark     & \ding{53}     & 74.8  & 63.1  & 68.1  & 55.6  & 67.1  & 65.7  \\
    CoWA  & ICML22  & \checkmark     & \ding{53}     & 80.6  & 69.0  & 67.2  & 60.0  & 69.0  & 69.2  \\
    PLUE  & CVPR23  & \checkmark     & \ding{53}     & 74.0  & 61.6  & 65.9  & 53.8  & 67.5  & 64.6  \\
    TPDS  & IJCV23  & \checkmark     & \ding{53}     & 77.1  & 66.4  & 67.0  & 58.2  & 68.6  & 67.5  \\
    \midrule
    DIFO-C-B32 & CVPR24 & \checkmark     & \checkmark     & 87.9  & 80.1  & 77.4  & 75.5  & 79.2  & 80.0  \\
    \rowcolor[rgb]{ .91,  .91,  .91} \textbf{DRIVE(Ours)} & -     & \textbf{\checkmark} & \textbf{\checkmark} & \textbf{88.1 } & \textbf{81.0 } & \textbf{78.2 } & \textbf{76.0 } & \textbf{79.8 } & \textbf{80.6 } \\
    \bottomrule
    \end{tabular}%
    }
  \label{tab:domainnet126}%
\end{table*}%



\subsection{Ablation Study}
To thoroughly evaluate the contributions of each component in DRIVE, we conducted an ablation study focusing on the following aspects: Entropy-Aware Predictor, Perturbed Model Encouragement, and Dynamic Perturbation Adjustment.

\noindent \textbf{Entropy-Aware Predictor.} The Entropy-Aware Predictor is designed to adjust label weights based on prediction uncertainty, ensuring the model focuses on reliable data while avoiding noisy regions. As shown in Table~\ref{tab:ablation}, enabling the Entropy-Aware Predictor alone (second row) improves performance on the A → D and D → W tasks, indicating its effectiveness in enhancing the model's robustness.

\noindent \textbf{Perturbed Model Encouragement.} The Perturbed Model Encouragement mechanism exposes one of the models to perturbations via projection gradient descent (PGD) guided by mutual information, focusing on high-uncertainty regions. When combined with the Entropy-Aware Predictor (third row), there is a further improvement in performance, particularly on the A → D and D → W tasks, demonstrating the importance of exploring the feature space more comprehensively.

\noindent \textbf{Dynamic Perturbation Adjustment.} The Dynamic Perturbation Adjustment mechanism dynamically adjusts the perturbation level based on the loss from the first stage, encouraging the model to explore a broader range of the target domain while preserving existing performance. Adding this component (fourth row) maintains the performance gains observed in the third row, indicating that dynamic adjustment is beneficial for maintaining robustness and generalization.

\begin{table*}[htbp]
  \centering
  \caption{Classification results of ablation study on several settings on Office31 (\%). The full results are provided in Supplementary.}
  \resizebox{1 \columnwidth}{!}{%
    \begin{tabular}{ccc|ccc}
    \toprule[0.2em]
    \multirow{2}[2]{*}{Entropy-Aware Predictor} & \multirow{2}[2]{*}{Perturbed Model Encouragement} & \multirow{2}[2]{*}{Dynamic Perturbation Adjustment} & \multicolumn{3}{c}{\textbf{Office31}} \\
          &       &       & A → D & A → W & D → W \\
    \midrule
    \ding{53}     & \ding{53}     & \ding{53}     & 96.39 & 95.72 & 95.6 \\
    \checkmark     & \ding{53}     & \ding{53}     & 96.99 & 95.72 & 95.97 \\
    \checkmark     & \checkmark     & \ding{53}     & 97.39 & 95.85 & 97.74 \\
    \checkmark     & \checkmark     & \checkmark     & 97.39 & 95.97 & 97.74 \\
    \bottomrule[0.2em]
    \end{tabular}%
  }
  \label{tab:ablation}%
\end{table*}

These findings highlight the importance of each component in DRIVE and validate the design choices made to ensure robust and effective domain adaptation. The combined use of the Entropy-Aware Predictor, Perturbed Model Encouragement, and Dynamic Perturbation Adjustment significantly enhances the model's performance across different domain adaptation tasks.

\subsection{Grad-CAM Visualization of SFUDA}

\begin{figure}
    \centering
    \includegraphics[width=0.9\columnwidth]{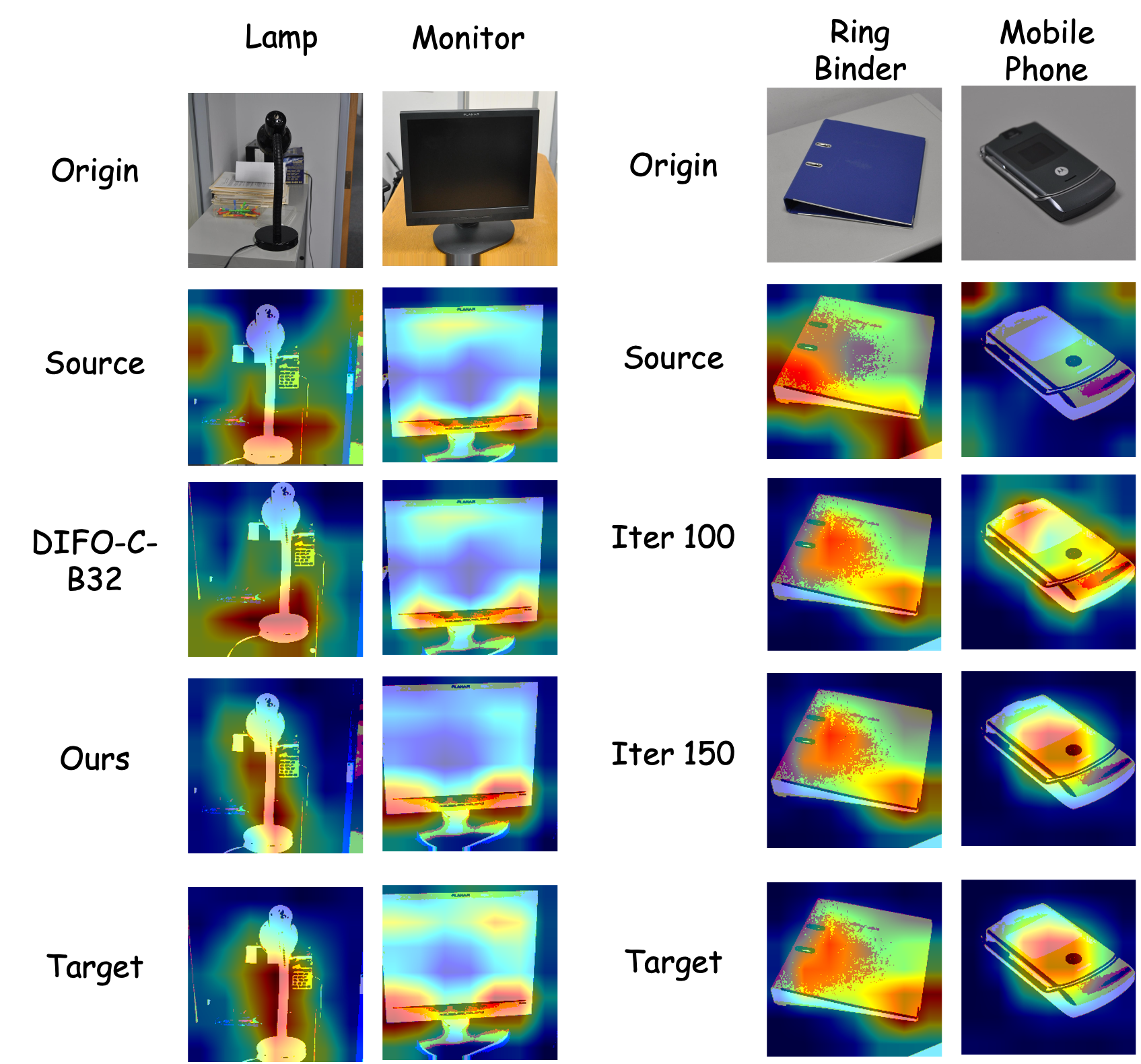}
    \caption{Grad-CAM visualization of our proposed method (DRIVE) compared to the primary baseline (DIFO), and the evolution of Grad-CAM visualizations for DRIVE as the number of iterations increases.}
    \label{fig:cam}
\end{figure}

To gain deeper insights into how DRIVE and competing models attend to domain-relevant features, we conducted Grad-CAM analysis on selected samples from the target domain, which shown on \ref{fig:cam}. By comparing the attention maps generated by DRIVE against those produced by competitive baseline(DIFO), we can assess the efficacy of our method in focusing on meaningful and domain-invariant features. The Grad-CAM visualizations thus provide empirical evidence supporting our claim that DRIVE not only achieves quantitative performance gains but also qualitatively attends to more meaningful and transferable features within the target domain. This characteristic is critical for achieving robust and reliable domain adaptation, especially in real-world applications where domain shifts are common and unpredictable.


\section{Conclusion}
Source-Free Unsupervised Domain Adaptation (SFUDA), involves adapting a pre-trained model to a target domain using only unlabeled target data, leading to issues such as overfitting, underfitting, and poor generalization due to domain discrepancies and noise. Existing SFUDA methods often struggle with these challenges due to their reliance on single-model architectures. To address these difficulties, we introduced DRIVE (Dual-Robustness through Information Variability and Entropy), a novel SFUDA framework leveraging a dual-model architecture. DRIVE captures diverse characteristics of the target domain by using two models initialized with identical weights, one of which is exposed to perturbations via projection gradient descent (PGD) guided by mutual information, focusing on high-uncertainty regions. Additionally, we introduced an entropy-aware pseudo-labeling strategy that adjusts label weights based on prediction uncertainty, ensuring the model focuses on reliable data while avoiding noisy regions. The adaptation process consists of two stages: the first aligns the models on stable features using a mutual information consistency loss, and the second dynamically adjusts the perturbation level based on the loss from the first stage, encouraging the model to explore a broader range of the target domain while preserving existing performance. This enhances generalization capabilities and robustness against interference. Evaluations on standard SFUDA benchmarks show that DRIVE consistently outperforms previous methods, delivering improved adaptation accuracy and stability across complex target domains.
\label{sec:con}


\bibliographystyle{unsrt}  
\bibliography{templateArxiv}

\clearpage

\end{document}